\documentclass{article}

\usepackage[preprint]{corl_2022} 

\usepackage{graphicx}  
\usepackage{amssymb}
\usepackage{amsmath}
\usepackage{stfloats}
\usepackage{caption}
\usepackage{subcaption}
\usepackage{wrapfig}

\title{Uncertainty estimation for Cross-dataset performance in Trajectory prediction}

\author{
  Thomas Gilles$^{1, 2}$, Stefano Sabatini$^{1}$, Dzmitry Tsishkou$^{1}$, Bogdan Stanciulescu$^{2}$, Fabien Moutarde$^{2}$\\
  $^{1}$Huawei Technologies France   $^{2}$ Mines ParisTech \\
  \texttt{thomas.gilles@mines-paristech.fr}
}

\begin{document}
\maketitle


\begin{abstract}
  While a lot of work has been carried on developing trajectory prediction methods, and various datasets have been proposed for benchmarking this task, little study has been done so far on the generalizability and the transferability of these methods across dataset. In this paper, we observe the performance of two of the latest state-of-the-art trajectory prediction methods across four different datasets (Argoverse, NuScenes, Interaction, Shifts). This analysis allows to gain some insights on the generalizability proprieties of most recent trajectory prediction models and to analyze which dataset is more representative of real driving scenes and therefore enables better transferability. Furthermore we present a novel method to estimate prediction uncertainty and show how it could be used to achieve better  performance across datasets.
  
\end{abstract}


\section{Introduction}

Trajectory prediction is an essential step of the autonomous driving pipeline, and needs to be robust to any situation the self-driving vehicle may encounter. A failure to correctly predict the future of a neighboring car may lead to dangerous situations and even collisions. However, as learned methods grow in performance and popularity \citep{liu2021survey, gomes2022review, karle2022scenario} by either extending existing traditional methods \citep{jouaber2021nnakf} or replacing them completely \citep{mercat2021motion},
so does the dependency to the data coverage these models are trained on. Such methods may encounter distributional shift due to changing geographical or weather conditions \citep{malinin2021shifts}. It becomes therefore crucial to study the adaptability and performance of these methods across varying distributions.

Multiple trajectory prediction datasets \citep{zhan2019interaction, chang2019argoverse, caesar2020nuscenes, malinin2021shifts} have been used separately to train and evaluate motion estimation models, but few works actually study the performance of their models on more than one of these datasets at a time, and even more importantly, no study has yet been done to evaluate the representative coverage and generalization potential of the datasets across each other.

One recurring proposal to help with distributional shift is the use of uncertainty estimation. However, while this uncertainty is presented as a shift detector \citep{malinin2021shifts, postnikov2021transformer, pustynnikov2021estimating}, no practical use of this value has been proposed yet to actually diminish the impact of distributional shift.

This work present two main contributions:
\begin{itemize}
    \item We realize the first cross-dataset study in vehicle trajectory prediction and assess which datasets transfer best to others.
    
    \item We introduce a new way of estimating the model uncertainty by training the prediction model to output an heatmap. The model uncertainty is measured evaluating the spread of the predicted heatmap. We demonstrate that using this uncertainty to control the diversity of the predicted future trajectories leads to better performance on both single and cross-dataset evaluation.
\end{itemize}

\section{Related work}

In order to be exhaustive, a trajectory prediction model needs to be multimodal so it can represents all possible futures. The common approach is to have the model predict $k$ possible trajectories and only train the one closest to ground truth \citep{cui2019multimodal, liang2020learning, narayanan2021divide}, or to use a mixture of Gaussians trained with their likelihood \citep{deo2018convolutional, mercat2020multi}. Variational methods can also be applied \citep{lee2017desire, rhinehart2018r2p2, tang2019multiple, mangalam2020not, salzmann2020trajectron++, alahi2016social, sadeghian2019sophie, casas2020implicit} to sample multiple outcomes. A more explicit approach is to leverage explicit existing modalities such as clusters \citep{phan2020covernet, chai2020multipath} or map elements \citep{zhao2020tnt, zhang2020map, zeng2021lanercnn, deo2021multimodal} to build possible trajectories upon. Transformer architectures also suit notably well to decode multiple modalities from separate learned embeddings \citep{yuan2021agentformer, girgis2021latent, ngiam2021scene,postnikov2021transformer, pustynnikov2021estimating}.
Another way to obtain an exhaustive outcome is to use a heatmap as the output of the model. This heatmap can either represent a single agent future distribution \citep{hong2019rules, kurbiel2020prognosenet, gilles2021home, gilles2021gohome, gu2021densetnt} or the occupancy of all present vehicles \citep{kim2017probabilistic, park2018sequence, schafer2022context, mahjourian2022occupancy}. 

Some methods use uncertainty-based losses to improve their prediction training \citep{ngiam2021scene, varadarajan2021multipath++} or take it as input to increase robustness to perception errors \cite{weng2021mtp, ivanovic2021propagating}, and a few apply it to predict distributional shift \citep{postnikov2021transformer, pustynnikov2021estimating}. But so far, apart from replacing high uncertainty cases with ground truth \citep{malinin2021shifts} for evaluation, little work has been done on how to actually leverage this uncertainty.

Recently, more reflection has been carried out on the ways of evaluating these trajectory prediction methods. Some argue that motion estimation should be evaluated with regards to its downstream effect on the planner \citep{ivanovic2021rethinking,ivanovic2021injecting, mcallister2022control}, while others focus on their lack of generalization to new scenarii \cite{bahari2022vehicle}.
Similar cross-datasets studies have been conducted for fields related to autonomous driving such as human intention \citep{gesnouin2022assessing} or detection \citep{hasan2022pedestrian}.

\section{Cross-dataset analysis in trajectory prediction}

In this first part of the paper, we focus on analyzing the cross-dataset performance of recent state-of-the-art trajectory prediction methods. We first define the trajectory prediction task and present two recent trajectory prediction methods that attack the problem from different perspectives. Subsequently, we describe the datasets we use in the analysis and finally we present the cross-dataset performance of both prediction methods. 

\subsection{Task definition}
Given a target agent, its past history and its surrounding context which consists of the neighbor agents and the road graph, the goal of a trajectory prediction model is to predict the future trajectory of the target agent up to a time horizon $T$. More precisely, we will simplify here the trajectory prediction problem to predicting the final destination point at the end of the prediction horizon $T$, as commonly done in \citep{zhao2020tnt, zeng2021lanercnn, gilles2021home, gu2021densetnt}. As the future is uncertain and can contain multiple possibilities, the prediction model needs to be multimodal and predict up to k=6 trajectory modalities, with matching probabilities.

\subsection{Trajectory prediction methods} \label{subsec:pred_methos}


In order to be representative of the wide scope of existing trajectory prediction methods, we implement two state-of-the-art baselines both representative of the different possibilities for output formulation, i.e. scalar coordinates output or probability heatmap output.

\subsubsection{SceneTransformer} 

SceneTransformer \citep{ngiam2021scene} is one of the most recent trajectory prediction model regressing multiple scalar trajectories using a transform architecture. In its encoding phase, it retains the time dimension across all agents present in the scene, and applies factorized self-attention either across agents or time, as well as cross-attention onto the map context. It uses modality one-hot embeddings and a transformer decoder to predict multiple modalities, so that it can share the decoding weights of the multiple futures that are training using a Winner-Take-All loss as in most scalar output methods \citep{cui2019multimodal, chai2020multipath, liang2020learning, narayanan2021divide, varadarajan2021multipath++}.

We reimplement a similar architecture with the same number of layers as in the original paper but with a smaller hidden dimension D=128 to make it fit on a single GPU and be more comparable to our second baseline in parameter size and training time.


\subsubsection{GOHOME} \label{subsec:gohome}

GOHOME \citep{gilles2021gohome} is part of a growing class of methods using occupancy grids \citep{kim2017probabilistic, park2018sequence, hong2019rules, kurbiel2020prognosenet, ridel2020scene, mangalam2020goals, gilles2021home, casas2021mp3, gu2021densetnt, gilles2021thomas, schafer2022context, mahjourian2022occupancy}. 
The occupancy grid usually represents a probability distribution in the form of a heatmap describing the possible future locations of the vehicle at the end of the prediction horizon $T$.
Given the predicted heatmap, a set of final future positions is sampled. In a final step, for each sampled locations, the full trajectory is regressed \citep{gilles2021gohome}. 
In order to sample the possible future locations from the heatmap, usually a Non-Maximum Suppression (NMS) method is applied \citep{gu2021densetnt, gilles2021home, schafer2022context}. This NMS requires a sampling radius parameter $r$ to determine how far apart the sampled endpoints should be from each other.

We apply some slight modifications to the GOHOME architecture to adapt it to our case analysis. First, since some datasets don't provide connectivity information between lanes \citep{malinin2021shifts}, we replace the graph convolutions with global attention, in a VectorNet-like manner as in \citep{gu2021densetnt, ngiam2021scene, pustynnikov2021estimating}. We also replace the lane-based heatmap decoder with the hierarchical sparse grid decoder from \citep{gilles2021thomas} for faster inference and once again independence from the HD-Map connectivity information

\subsection{Datasets and Metrics}

We evaluate performance on the widely used trajectory datasets Argoverse \citep{chang2019argoverse}, Interaction \citep{zhan2019interaction}, NuScenes \citep{caesar2020nuscenes} and Shifts \citep{malinin2021shifts}, all focusing on car trajectories. These benchmarks have slightly different initial settings as described in Tab. \ref{tab:data}. Namely, the history and prediction horizons are not always the same, and can be sampled at different rates. For fair evaluation and transferability, we standardize these datasets to always use 1s of history and predict 3s in the future. We also interpolate the trajectories to resample them at 10Hz each. 

\begin{table}[t!]
\caption{Dataset settings}
    \begin{center}
    \begin{tabular}{l|c c c c}
      \hline
      Dataset & Argoverse & Interaction & NuScenes & Shifts \\
      \hline
      History (s)  &  2 & 1 & 2 & 5\\
      Prediction horizon (s)   &  3 & 3 & 6 & 5\\
      Frequency (Hz) & 10 & 10 & 2 & 5 \\
      Training size & 200k & 400k & 30k & 5M\\
      \hline
    \end{tabular}
    \end{center}
    \label{tab:data}
\end{table}


In our analysis we consider the well established multimodal metrics minFDE$_l$ and MR$_l$ \citep{zhan2019interaction, chang2019argoverse}. minFDE$_l$ represents the minimum final displacement error at time horizon $T$ over the $l$ top-ranked trajectories. MR$_l$ represents the percentage of samples in the dataset on which the ground truth future position of the target agent at time horizon $T$ is farther than 2m from any of the $l$ top-ranked predicted trajectories.





\begin{figure}
\begin{subfigure}{.5\textwidth}
  \centering
  \includegraphics[width=.99\linewidth]{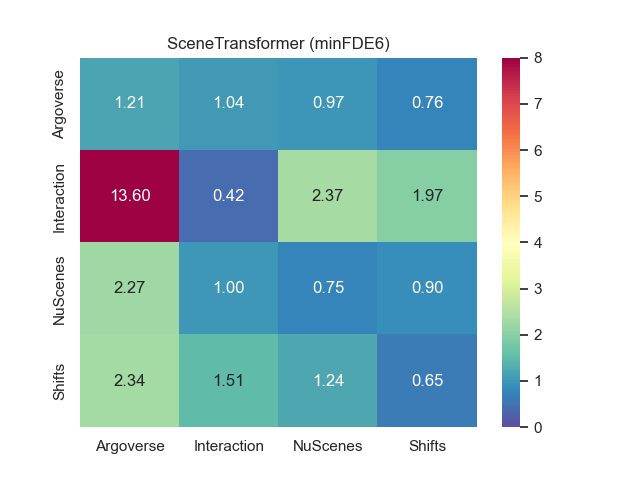}  
  \label{fig:sub-first}
\end{subfigure}
\begin{subfigure}{.5\textwidth}
  \centering
  \includegraphics[width=.99\linewidth]{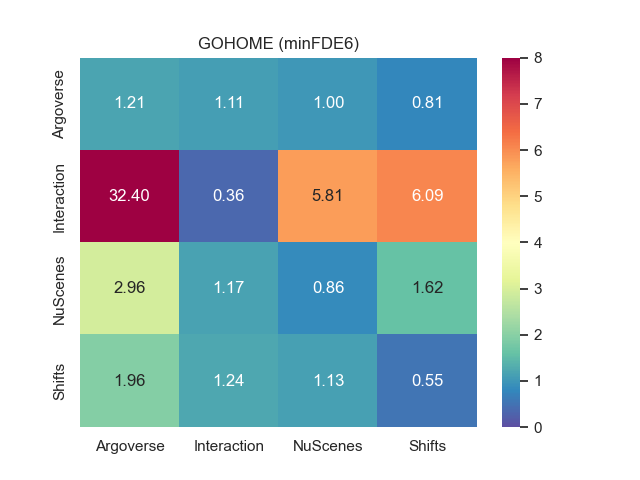}  
  \label{fig:sub-second}
\end{subfigure}


\begin{subfigure}{.5\textwidth}
  \centering
  \includegraphics[width=.99\linewidth]{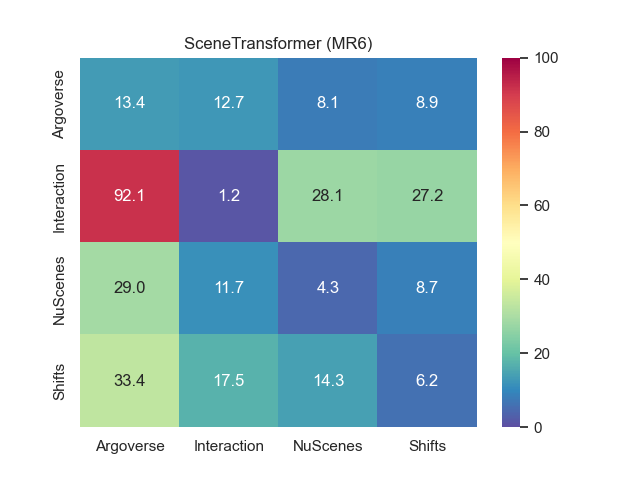}  
  \label{fig:sub-third}
\end{subfigure}
\begin{subfigure}{.5\textwidth}
  \centering
  \includegraphics[width=.99\linewidth]{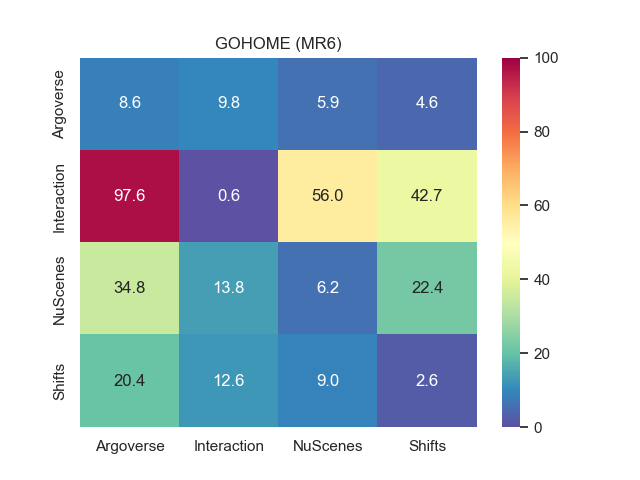}  
  \label{fig:sub-fourth}
\end{subfigure}
\caption{Prediction performance in a cross-datasets evaluation setting.}
\label{fig:cross_dataset_results}
\end{figure}


\subsection{Cross-dataset evaluation}
We analyze here the trajectory prediction performance of both models presented in section \ref{subsec:pred_methos} when they are trained on the training split of one dataset and tested on the validation splits of all datasets.

\subsubsection{Training details}

We trained each model for 50 epochs of 2000 iterations each, with a batch size of 64. The GOHOME hyper-parameter $r$ related to the sampling radius has been optimized on the training-split of the training dataset and kept unchanged for the test datasets.  
Few data augmentation schemes were employed to optimize the generalizability performance. First, all models are trained with random rotations to prevent overfitting on the current car heading measurement. Furthermore, we noticed that the Argoverse dataset does not present any case where the target agent to be predicted has a speed lower than 1 m/s. Contrarily, other datasets include vehicles to be predicted that stand still or with very low speed values for the whole prediction horizon. For this reason, it has been necessary to augment the Argoverse dataset with prediction samples related to vehicles present in the scenes that move slowly or are parked other than the predefined target. Without this augmentation procedure, models trained on the plain Argoverse ends up with poor generalization performance on other datasets. 


\subsubsection{Results}

We report onto Fig. \ref{fig:cross_dataset_results} the cross-dataset performance matrices for minFDE$_6$ and MR$_6$ for both prediction models. The label on the rows indicates the dataset used for training while the label on the columns represent the target test datasets. The numbers in the matrix corresponds to the performance measured on the validation split of the corresponding target dataset. As expected, the best performance are visible on the diagonals, since both models perform better when tested on data coming from the same distribution of the training.

We observe that  Argoverse training exhibits the smallest loss of performance when tested on other datasets. We also observe that despite its relatively short size (only 30k samples), the training on NuScenes also performs reasonably on other datasets, whereas when trained on Interaction, the models performs poorly on every other domain. We attribute the poor performance of Interaction to its different data collection and processing, done with top-view images from drones instead of the usual perception pipeline from the autonomous vehicle. Therefore Interaction is trained on an almost perfect object detection and tracking, and does poorly on other datasets filled with detection inaccuracies and tracking jumps caused by occlusions. Surprisingly, despite its superior sample size, training on Shifts doesn't provide better transferability performance compared Argoverse and NuScenes.

In order to assess ideal cross-dataset performance, for completeness we also present the results that are obtained when training the models on all the available datasets at the same time. To achieve this, each sample loaded during training is drawn randomly from one of the 4 datasets, with equal probability.
The reader is referred to the supplementary material for this analysis.

The main first conclusion we can draw from this cross-dataset performance is that it is not so much the size of the data that matters, rather than its ability to faithfully represent real conditions.

When comparing the performance between the heatmap-based and the scalar-based models, we can notice how the heatmap output provides the best MR on the training datasets (with the exception of NuScenes) while scalar output provide the best minFDE. Regarding the transferability performance, SceneTransformer present the smallest performance loss compared to GOHOME when tested on other datasets.

\begin{figure}[t!]
    \center
    \begin{subfigure}[b]{0.50\textwidth}
        \includegraphics[width=\textwidth]{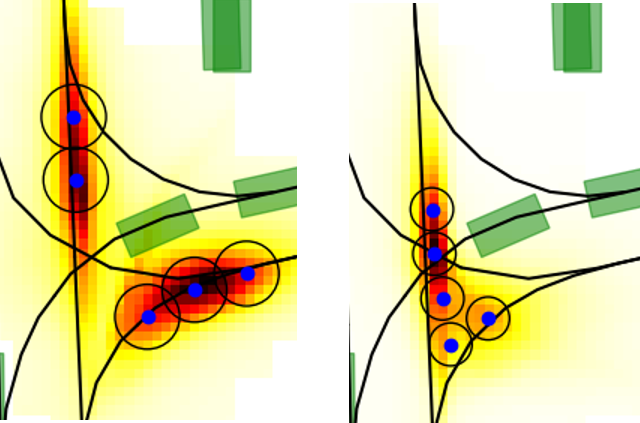}
        \label{fig:rad_sampl}
    \end{subfigure}
    \begin{subfigure}[b]{0.45\textwidth}
        \includegraphics[width=\textwidth]{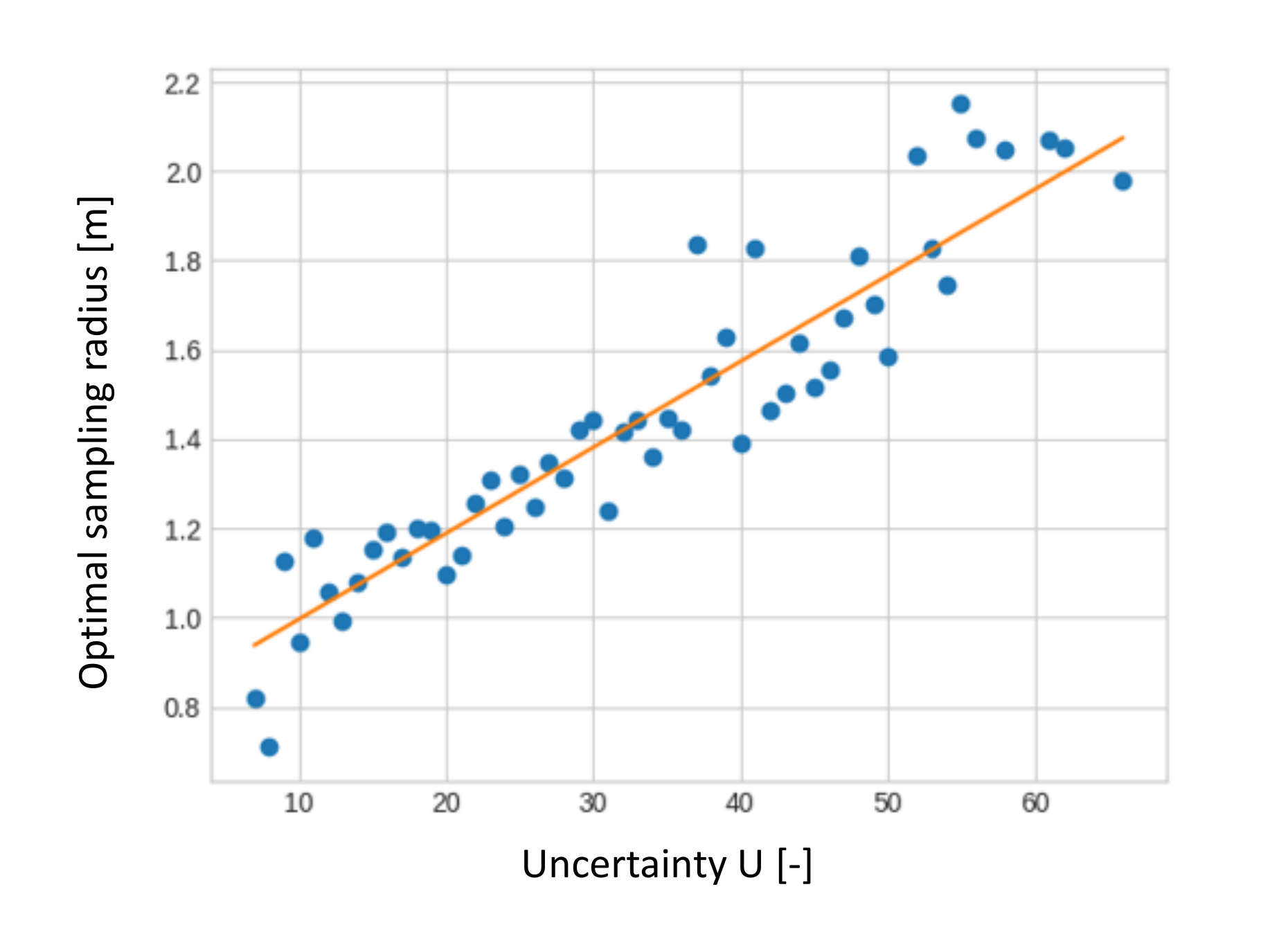}
        \label{fig:opt_radius}
    \end{subfigure}
    \caption{\textbf{On the left}: an example of sampling final locations with a different radius depending on the uncertainty of the heatmap. \textbf{On the right}: Optimal sampling radius for each value of the estimated uncertainty.}
    \label{fig:rad_sampl}
\end{figure}

\section{Heatmap-Based Uncertainty estimation}

In this section, we present a method to leverage the heatmap output formulation of models like GOHOME in order to estimate how much the model is uncertain when performing a trajectory prediction. We first present the formulation of the uncertainty estimation and in a second step we show how the uncertainty can be utilized to improve prediction performance.

\subsection{Uncertainty formulation}

The proposed uncertainty formulation is based on the fact that prediction methods designed to produce an heatmap provide a natural intrinsic uncertainty estimator in the spread of their output. We use the variance of the predicted spatial probability distribution as an indicator of the model uncertainty $U$: 


\begin{equation}
U = \sum_p H(p) \rVert p-E \rVert^2
   \quad\text{with}\quad 
E = \sum_p H(p) p \\
\label{eq:uncertanty_defintion}
\end{equation}

where we iterate over the positions $p$ of the heatmap and we indicate with $H(p)$ the probability value for the given position. $E$ corresponds to the expected value of the probability distribution described by the heatmap. With this formulation, we claim that the heatmap provides for free an unconstrained and non-parametric measure of uncertainty without the need of adding and training a model part specific to uncertainty prediction as in \citep{pustynnikov2021estimating} and \citep{postnikov2021transformer}.



\subsection{Controlling prediction diversity with uncertainty}

We leverage the presented uncertainty estimation to control the diversity of the predicted future locations at the prediction horizon $T$.  Intuitively, when the network is more uncertain, in order to minimize the  prediction error it is required to increase the diversity of the predictions to cover a wider span of possibilities. 
In practice we control this behavior, by adapting the sampling radius $r$ presented in section  \ref{subsec:gohome} to adjust the diversity of the sampled locations to the spread of the heatmap. This behavior is depicted in Fig.  \ref{fig:rad_sampl} where a bigger sampling radius is employed on the left example to cope with a more uncertain prediction testified by the bigger spread in the heatmap.



\subsection{Results}


\subsubsection{Uncertainty as prediction error estimator}

\begin{figure*}[t!]
\centerline{\includegraphics[width=1.\columnwidth]{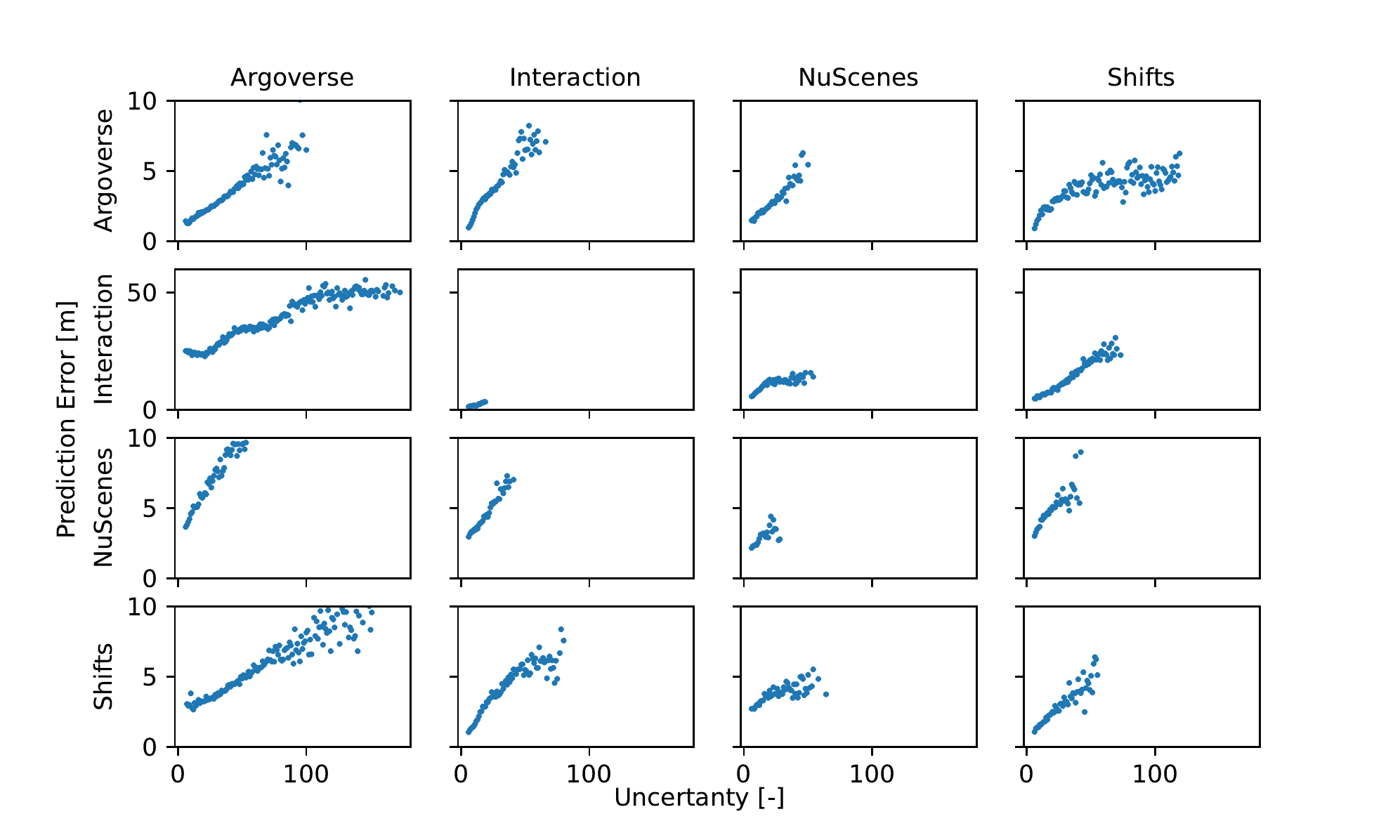}}
\caption{Analysis of correlation between uncertainty and prediction error in a cross-dataset setting.}
\label{fig:uncertanty_error}
\end{figure*}

To motivate our uncertainty definition, we first show how the uncertainty estimated through Equation \ref{eq:uncertanty_defintion} is correlated with the prediction error that the model ends up making. Figure \ref{fig:uncertanty_error} shows the average prediction error minFDE$_{1}$ of the GOHOME model for uncertainty values grouped into integer bins. In the case of GOHOME, minFDE$_{1}$ represents the error made by a single prediction on the most probable location highlighted by the heatmap. The prediction error is calculated on the validation split of each dataset in consideration. 
We can clearly see a strong correlation between uncertainty and prediction error testifying that heatmap based methods intrinsically carry a notion of their performance when making a prediction inference. It is interesting to notice how a similar trend is mantained even when the analysis is done cross-dataset, i.e. when the model is evaluated on a dataset different from the training dataset. 

\subsubsection{Uncertainty enhanced cross-dataset performance}

In this section we show the benefit of using the presented uncertainty to adapt the diversity of the predicted locations through the sampling radius $r$. 
First, we experimentally show that the optimal sampling radius $r_{opt}$ that minimize the prediction error follows a linear trend with respect the estimated uncertainty. Figure \ref{fig:rad_sampl} depicts the average optimal sampling radius calculated over the Argoverse dataset versus the estimated uncertainty grouped in bins of integer values. The reader is invited to check the supplementary materials for the plots on other datasets. 

\begin{figure*}[t]
    \center
    \begin{minipage}[t]{0.32\textwidth}
        \includegraphics[width=\textwidth]{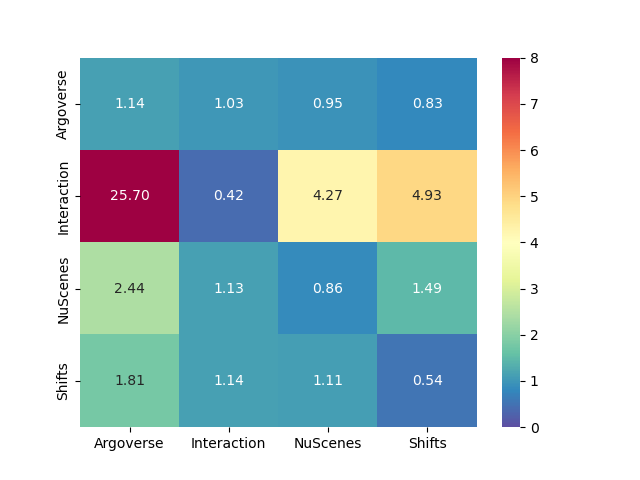}
        \label{fig:cross_var}
    \end{minipage}
    \begin{minipage}[t]{0.32\textwidth}
        \includegraphics[width=\textwidth]{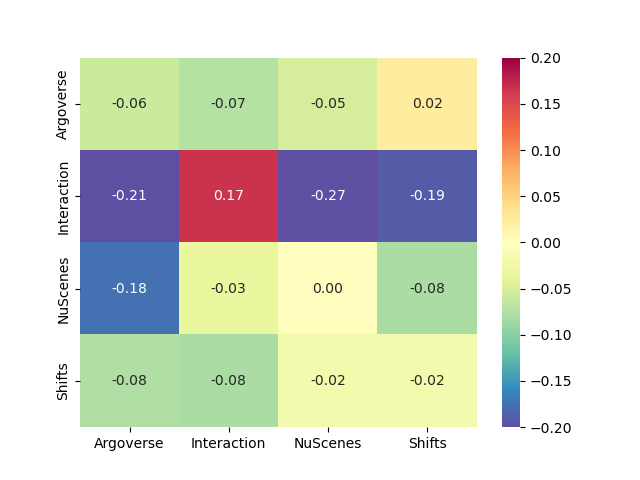}

        \label{fig:cross_radius}
    \end{minipage}
        \begin{minipage}[t]{0.32\textwidth}
        \includegraphics[width=\textwidth]{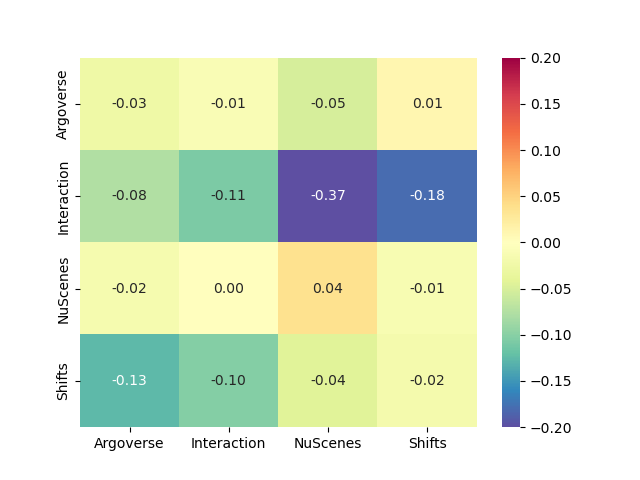}

        \label{fig:cross_radius}
    \end{minipage}
    \caption{\textbf{On the Left:} Absolute cross-dataset performance when using a sampling strategy based on uncertainty \textbf{On the Middle:} Relative improvement in cross-dataset minFDE$_{6}$ when using uncertainty compared to fixed radius sampling \textbf{On the right:} Relative improvement in cross-dataset minFDE$_{6}$ when using our heatmap-based uncertainty compared to a learned uncertainty baseline}
    \label{fig:adaptive_radius_matrix}
\end{figure*}

Furthermore, we report cross-datasets results using an adaptive sampling radius to adjust the prediction diversity depending on uncertainty. Left image in Fig. \ref{fig:adaptive_radius_matrix} shows the minFDE$_{6}$ cross-dataset performance when the model is trained on the dataset denoted in the row label and evaluated on the datasets denoted by the column label. In each one of this experiment the radius is adapted following the linear model calibrated on the dataset used for training and kept unchanged for evaluation on target datasets. We can see in the middle image of \ref{fig:adaptive_radius_matrix} how the adaptive sampling strategy is significantly better in almost all cases compared to the constant radius sampling presented in Fig. \ref{fig:cross_dataset_results}.



We benchmark our method of computing the prediction uncertainty by comparing it to a learned variance $V$ of a Gaussian distribution as in \citep{kendall2017geometric, meyer2020learning, ngiam2021scene, moreau2022coordinet}. As in \citep{kendall2017geometric} we directly predict $s=\log(V)$ for numerical stability with the following loss:

\begin{equation}
L(s) = E \cdot exp(-s) + s
   \quad\text{with}\quad 
E = minFDE_{6} \\
\label{eq:uncertanty_baseline_los}
\end{equation}

\begin{figure*}[h]
\centerline{\includegraphics[width=1.\columnwidth]{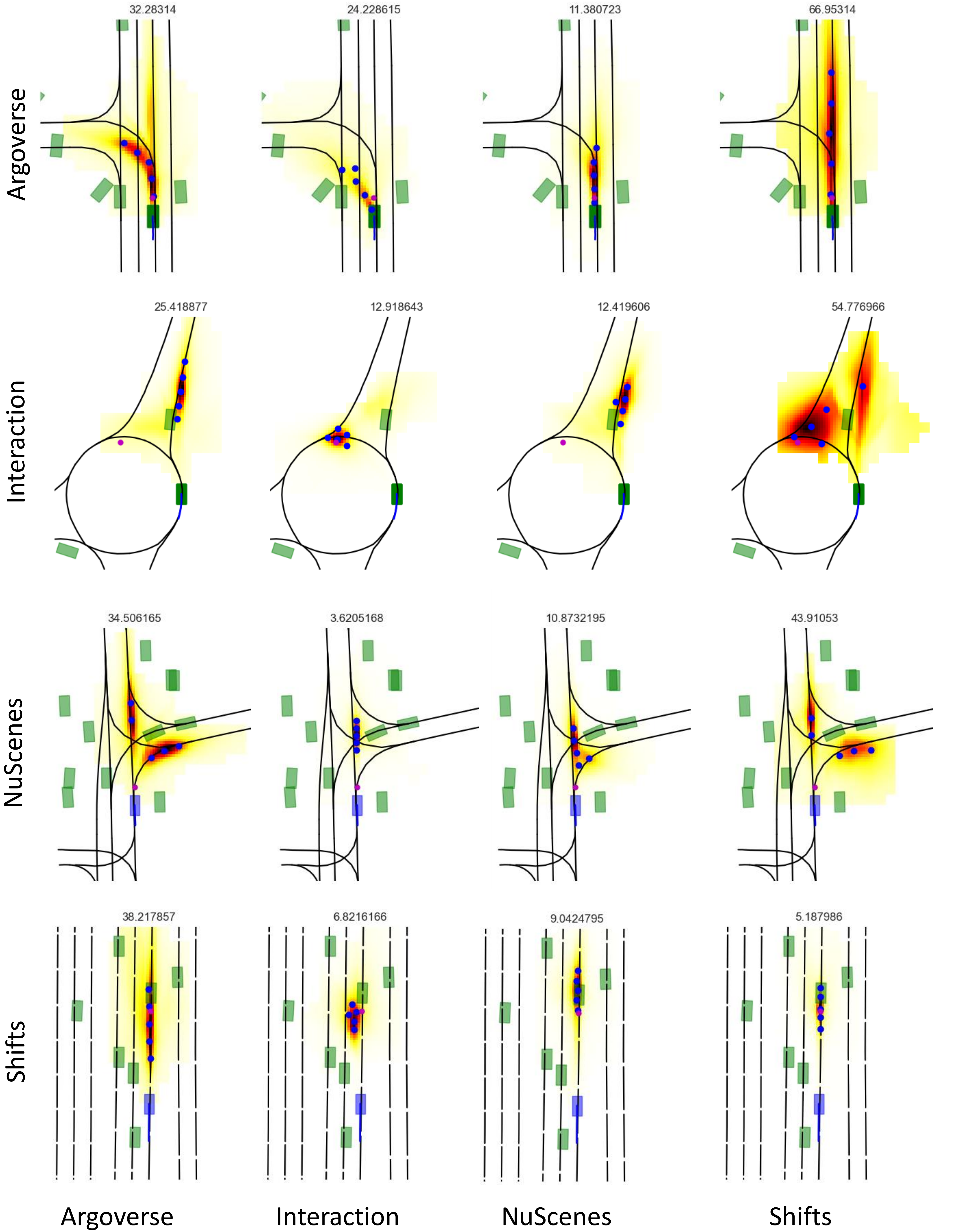}}
\caption{Qualitative results across datasets. Sampled endpoints are displayed in blue and ground truth in magenta. Heatmap variance is displayed on top of each example}
\label{fig:quali}
\end{figure*}

We report on the right image of Fig. \ref{fig:adaptive_radius_matrix}, the improvement in minFDE$_{6}$ when using our uncertanty definition of Equation \ref{eq:uncertanty_defintion} compared with the learned baseline of Equation \ref{eq:uncertanty_baseline_los} to adapt the sampling radius. While it yields similar results on the same train-test diagonal, the learned uncertainty tends to overfit on its training data and doesn't perform as well on out-of-distribution data.
We display qualitative prediction examples in Fig. \ref{fig:quali}. Each line is an example sample of one dataset, and each column the prediction result of the model trained on the corresponding dataset. We also report uncertainty numbers for each sample to observe how uncertainty matches the heatmap spread and the resulting adapted endpoint sampling


%

\section{Conclusion}
In this work we have done the first cross-dataset analysis in the field of vehicle trajectory prediction for autonomous driving. We have analyzed the cross-dataset transferability performance of two state of the art trajectory prediction model. We have also proposed a new way to estimate uncertainty for heatmap-based trajectory prediction methods that doesn't require any further training and works better than classically learned uncertainties. We showed how using the uncertainty boost trajectory prediction performance of heatmap-based methods in a cross-dataset setting.

\section{Weaknesses}
This analysis has been limited to car trajectory prediction, a similar analyses across different type of traffic participants such as bicycles and pedestrians would also be of interest.
Furhermore, while this study demonstrates that the use of heatmap variance for uncertainty estimation and sampling radius adaptation brings a significant performance improvement to heatmap output methods, the comparison to scalar outputs methods like SceneTransformer shows a less clear trend. The heatmap base method enriched with uncertainty has similar transferability performance to SceneTransformer. SceneTransformer is trained end-to-end to directly predict a set of multimodal coordinates and somehow internally learns to adapt the diversity of the predictions without explicitly outputting an uncertainty value. On the other hand we strongly believe that having an explicit uncertainty output can be useful also for other downstream tasks in the autonomous driving stack.




\bibliography{example}  

\clearpage

\appendix

\section{Comparison of data distribution between datasets and influence of data augmentation}

\subsection{Speed distribution}

\label{sec:distrib}

\begin{figure*}[h]
\centerline{\includegraphics[width=1.\columnwidth]{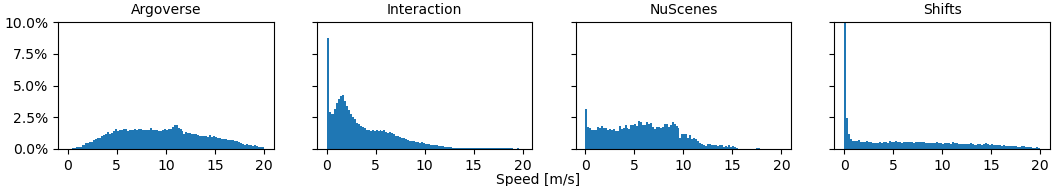}}
\caption{Distribution of average speed between initial agent position and last future position. Shifts reaches more than 40$\%$ samples of 0 m/s average speed, this bin is therefore out of scale for easier cross-dataset comparison.}
\label{fig:speed_distrib}
\end{figure*}

We mention in Sec. 3.4.1 the need to include non-target agents in the training data of Argoverse to correctly generalize to other datasets. We illustrate here this distributional gap in Fig. \ref{fig:speed_distrib}, where we display the average speed of the target agent during the future to be predicted. We can therefore observe than Argoverse has little to no agent that stay stationary, compared to other datasets

\begin{figure}[h]
    \centering
    \begin{subfigure}[b]{0.4\textwidth}
        \centering
        \includegraphics[width=\textwidth]{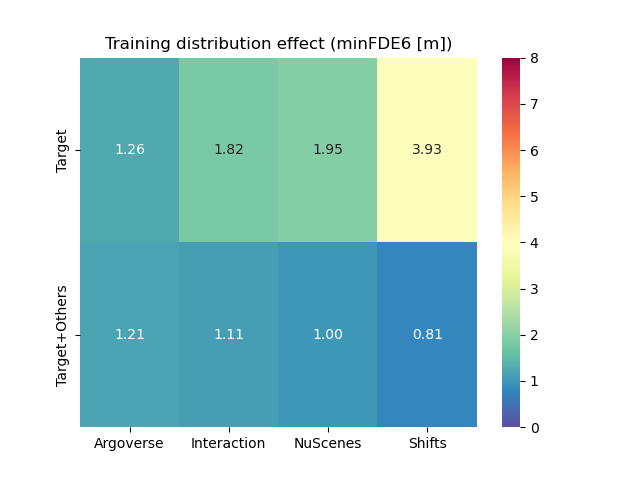}
        \caption{Performance difference}
        \label{fig:argo_perf}
    \end{subfigure}%
    ~
    \begin{subfigure}[b]{0.5\textwidth}
        \centering
        \includegraphics[width=\textwidth]{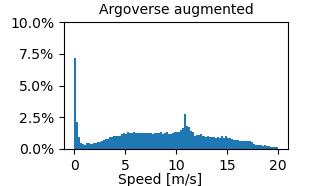}
        \caption{Speed distribution after augmentation}
        \label{fig:argo_speed}
    \end{subfigure}
    \caption{Impact of incorporating non-target agents in Argoverse to demonstrate slow-moving behaviors}
    \label{fig:rad_sampl}
\end{figure}

This leads to the performance gap observed in Fig. \ref{fig:argo_perf}, where the model trained strictly on Argoverse has way higher errors on the other datasets, and notably on the Shifts dataset which has a very high proportion of stationary samples. However, when we include a random sampling of 30$\%$ of agents other than the predefined target, the resulting speed distribution reported in Fig. \ref{fig:argo_speed} is much more representative of lower speed cases, and transfers much better onto the other datasets without losing performance on Argoverse itself.

\subsection{Noise distribution}
\label{sec:distrib_noise}

\begin{figure}[h]
\centerline{\includegraphics[width=1.\columnwidth]{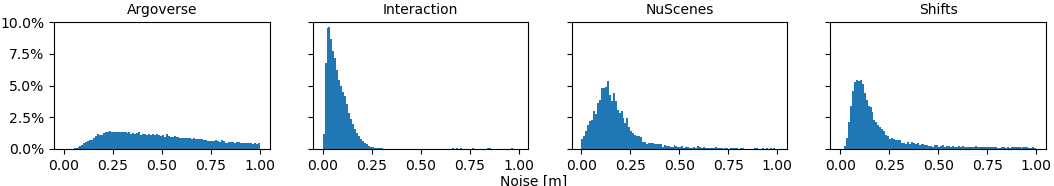}}
\caption{Distribution of perception noise across each dataset}
\label{fig:noise_distrib}
\end{figure}

In order to estimate perception noise in each dataset, we filter each trajectory with a Kalman filter and report the maximum displacement between the raw trajectory and the filtered one. We report the resulting noise distribution in Fig. \ref{fig:noise_distrib} and notice that the Interaction distribution is shifted towards lower noises than the other datasets, while Argoverse reaches higher noise values. These differences may explain the poor performance of the Interaction-trained model on other datasets.

\section{Adaptative radius for endpoint sampling}

GOHOME outputs a heatmap estimating the probability distribution of the position of the target agent, onto which we apply a Non-maximum Suppression (NMS) method to sample the desired number of endpoint modalities. This NMS requires a sampling radius parameter $r$ to determine how far apart the sampled endpoints should be from each other. We illustrate in Fig. \ref{fig:rad_sampl} the effect of this radius on the sampling.

\begin{figure*}[h]
\centerline{\includegraphics[width=1.\columnwidth]{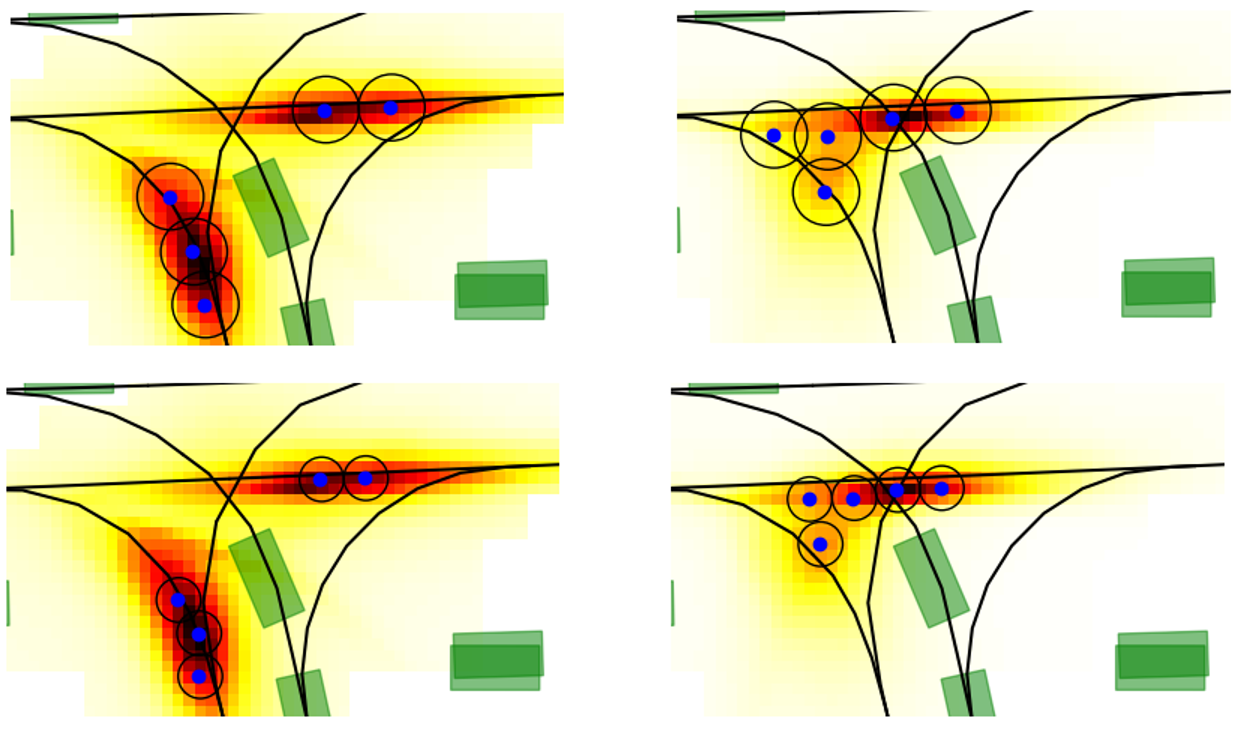}}
\caption{\textbf{Left column}: High uncertainty heatmap. \textbf{Right column}: Low uncertainty heatmap. \textbf{Top line}: high sampling radius. \textbf{Bottom line}: low sampling radius. As seen in the bottom left, using a low radius for a very spread heatmap leads to uncovered areas that may account for missed predictions. On the other hand, setting a high radius on a very focused heatmap spreads the sampled endpoints more than necessary and may generate a higher error if the ground truth is in-between two sampled points.}
\label{fig:rad_sampl}
\end{figure*}

\begin{figure*}[t]
\centerline{\includegraphics[width=1.\columnwidth]{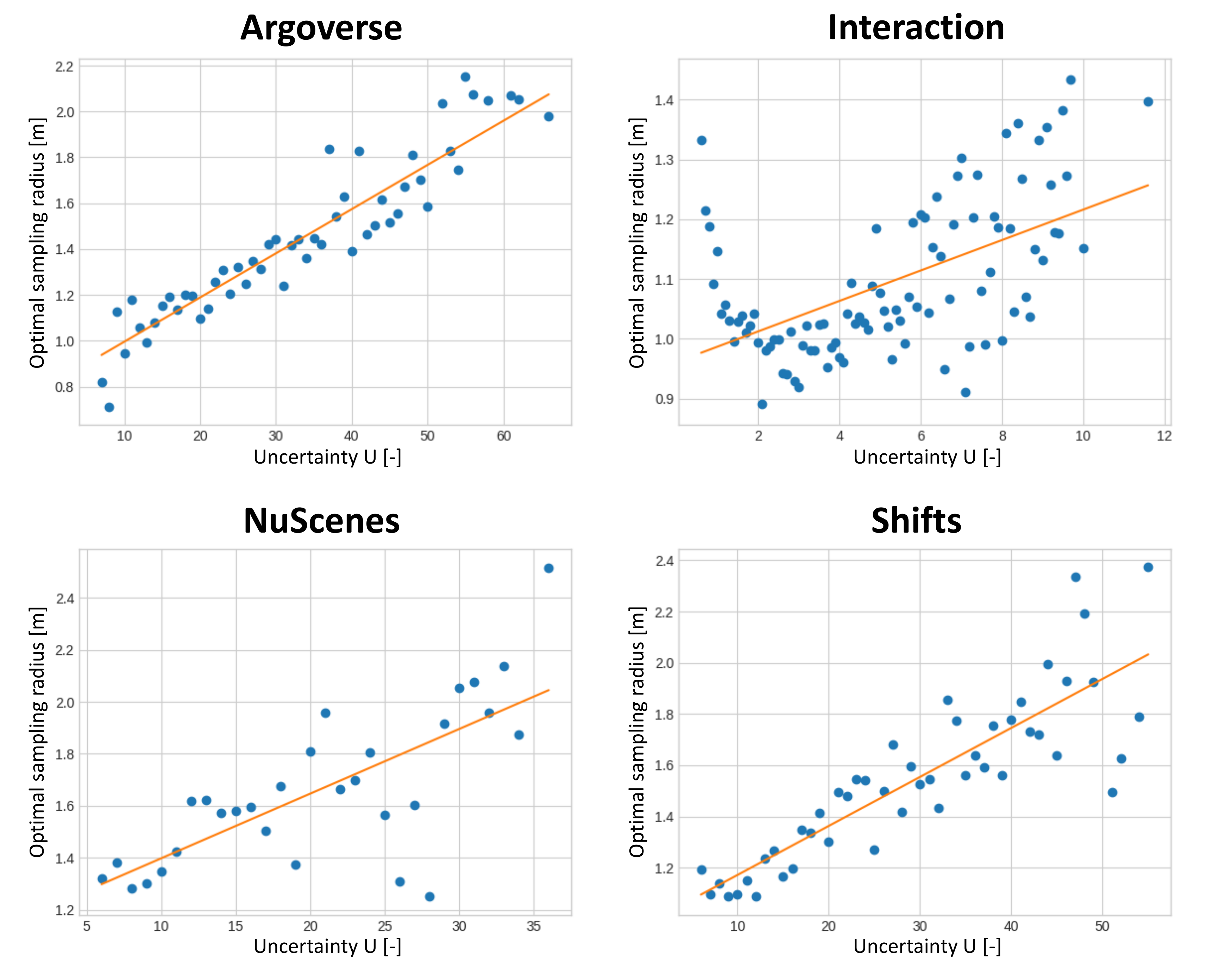}}
\caption{Average optimal radius with regard to uncertainty. We bin uncertainty values per equal integer values, and average the optimal radius for each of the cases in the bin. We plot in orange the linear curve obtained from applying least square regression on the points. }
\label{fig:radius_plot}
\end{figure*}

As seen in the Fig. \ref{fig:rad_sampl} above, given a fixed number of future modalities, the distance between these future points should be adapted with regards to how spread the heatmap is, which correlates with the uncertainty of the model. We demonstrate this correlation further in Fig. \ref{fig:radius_plot}, where we plot for each dataset, for the model trained on this dataset, the average optimal radius (according to the minFDE$_6$ metric) for uncertainty values grouped into integer bins. 

Fig. \ref{fig:radius_plot} highlights that this intuition for adaptive sampling is present on a per-sample scale, for most of the uncertainty values range. We therefore apply ordinary least squares to find regression coefficients between our estimated uncertainty and the optimal radius for a given case. The resulting curves are plotted in Fig. \ref{fig:radius_plot} and we report the resulting regression coefficients in Tab. \ref{tab:grid_radius}, as well as the optimal fixed radii without adaptation for each dataset.

\begin{table}[h]
\caption{Optimal radii and linear regression parameters per dataset}
    \begin{center}
        \begin{tabular}{l|c c c c}
          \hline
          Dataset & Argoverse & Interaction & NuScenes & Shifts \\
          \hline
          Radius  &  1.5 & 0.6 & 1.1 & 1.5\\
          Affine   &  0.020x+0.78 & 0.026x+0.96 & 0.014+1.32& 0.022x+0.91\\
          \hline
        \end{tabular}
    \end{center}
    \label{tab:grid_radius}
\end{table}

\section{Performance from training on all datasets}

To get a better estimation of what ideal generalization cross-dataset performance would be, we train a model on all datasets at the same time, each sample being drawn from one of the dataset with equal 25$\%$ probability. We report in Tab. \ref{tab:mixed_training} the results of both GOHOME (with fixed or adaptive radius) and SceneTransformer models trained in this setting.

\begin{table}[h]
\caption{Prediction performance minFDE$_{6}$ in a mixed dataset training setting }
    \begin{center}
    \begin{tabular}{l|c c c c}
      & Argoverse & Interaction & NuScenes & Shifts \\
     \hline
     GOHOME (fixed r=1.5m) & 1.34 & 0.66 & 0.88 & 0.70 \\
     GOHOME (adaptive radius) & 1.24 & 0.63 & 0.85 & 0.66 \\
     SceneTransformer & 1.33 & 0.58 & 0.81 & 0.58 \\
     \hline
    \end{tabular}
    \end{center}
    \label{tab:mixed_training}
\end{table}

\begin{figure}[h]
\centerline{\includegraphics[width=1.\columnwidth]{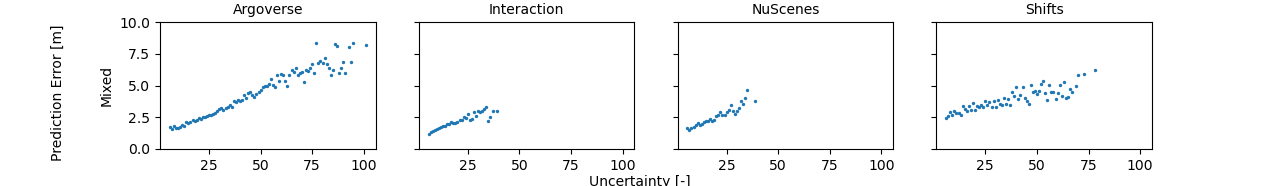}}
\caption{Analysis of correlation between uncertainty and prediction error across datasets for a model jointly trained on a mix of all datasets.}
\label{fig:mixed_error}
\end{figure}

We show in Fig. \ref{fig:mixed_error} the average prediction error with regards to the predicted uncertainty per data point for the GOHOME model. Compared to the single-dataset trained models, we observe that, while having lower variance, the error curves show similar trends when trained on all datasets and reach  similar ranges.

\begin{figure}[h]
\centerline{\includegraphics[width=1.\columnwidth]{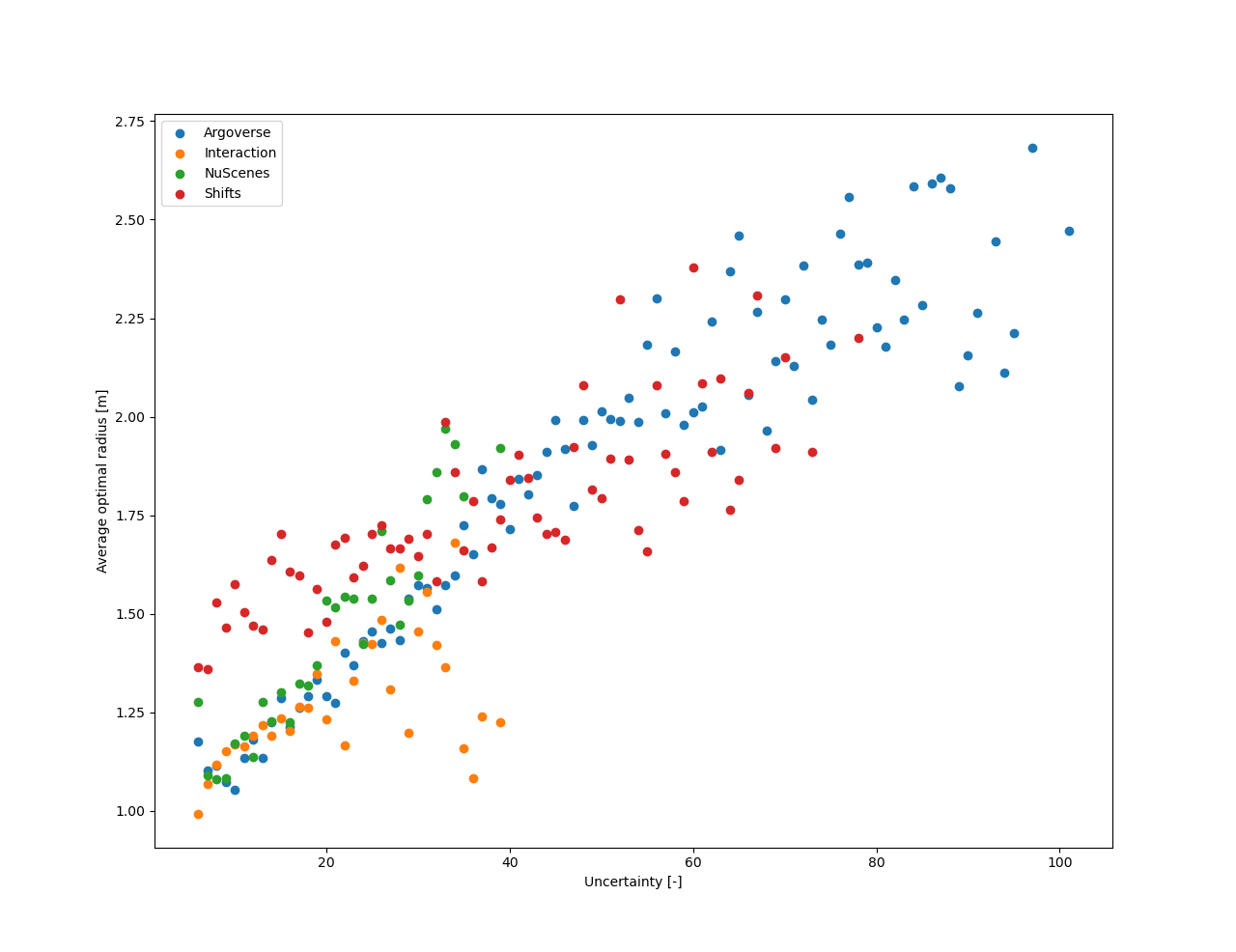}}
\caption{Optimal sampling radius for each value of the estimated uncertainty for a model trained jointly on a mix of all datasets.}
\label{fig:mixed_radius}
\end{figure}

We also report the optimal sampling radius of the GOHOME model with regards to uncertainty in Fig. \ref{fig:mixed_radius}. We highlight the data points from different datasets in different colors, and notice that each dataset has slightly different ranges but similar linear correlation trends.

\section{Unsuccessful trials}

Following the noise observations made in Sec. \ref{sec:distrib_noise}, we tried to augment the training data on Interaction with synthetic perception noise to bridge the gap to the other datasets. We were however not able to gain any kind of significant performance this way. This failure may be due to the way we modelled perception noise (independent Gaussian noise at every timestep) that could be inappropriate, or to the fact that the performance gap is due to other factors other than input noise.

We also noticed a difference in speed distribution in Fig. \ref{fig:speed_distrib} that reaches a lower upper limit (approx. 12.5 m/s) in Interaction compared to other datasets (although NuScenes also has a similarly limited distribution), and tried global random scaling to simulate higher speed, but this didn't bring much improvement either.

We hypothesize that the remaining performance gap when training on Interaction may be due to overfitting on the limited number of maps, as Interaction has a discrete set of relatively small intersection maps compared to other dataset maps that scale closer to city sizes, but didn't explore this hypothesis further.

\end{document}